\documentclass[runningheads]{llncs}
\usepackage{subcaption}
\usepackage{float}
\usepackage{amssymb}
\usepackage{threeparttable}

\usepackage{algorithmic}
\usepackage[ruled,vlined]{algorithm2e}
\usepackage{amsmath}
\usepackage{algorithmic}
\usepackage{comment}
\usepackage{booktabs}
\usepackage{cite}
\usepackage{multirow}
\usepackage{listings}
\usepackage{bibentry}
\usepackage{caption}
\usepackage{wrapfig}
\usepackage{newfloat}
\usepackage{amsfonts}
\usepackage{graphicx}
\usepackage{csquotes}
\usepackage{bbding}

\title{Two-stage Denoising Diffusion Model for Source Localization in Graph Inverse Problems}

\author{Bosong Huang\inst{1} \and
Weihao Yu\inst{2}\and
Ruzhong Xie\inst{1} \and
Jing Xiao\inst{1} \and
Jin Huang (\Envelope) \inst{1}
}
\authorrunning{F. Author et al.}
\institute{South China Normal University, Guangzhou, China \email{\{bosonghuang,rzxie,xiaojing,huangjin\}@scnu.edu.cn} \and
Research Institute of China Telecom Corporate Ltd., Guangzhou, China
\email{yuwh3@chinatelecom.cn}}
\begin{document}
\maketitle

\begin{abstract}
  Source localization is the inverse problem of graph information dissemination (information diffusion) and has broad practical applications. 
  However, the inherent intricacy and uncertainty in information dissemination pose significant challenges, and the ill-posed nature of the source localization problem further exacerbates these challenges. Recently, deep generative models, particularly diffusion models inspired by classical non-equilibrium thermodynamics, have made significant progress. While diffusion models have proven to be powerful in solving inverse problems and producing high-quality reconstructions, applying them directly to the source localization problem is infeasible for two reasons. Firstly, it is impossible to calculate the posterior disseminated results on a large-scale network for iterative denoising sampling, which would incur enormous computational costs. Secondly, in the existing methods designed for this field, the training data itself are ill-posed (many-to-one); thus simply transferring the diffusion model would only lead to local optima.
  To address these challenges, we propose a two-stage optimization framework, the source localization denoising diffusion model (SL-Diff). In the coarse stage, we devise the source proximity degrees as the  supervised signals to generate  coarse-grained source predictions. This aims to efficiently initialize the next stage, significantly reducing its convergence time and calibrating the convergence process. Furthermore, the introduction of cascade temporal information in this training method transforms the many-to-one mapping relationship into a one-to-one relationship, perfectly addressing the ill-posed problem. In the fine stage, we design a diffusion model for the graph inverse problem that can quantify the uncertainty in the dissemination process. Thanks to the excellent collaboration of the two stages, the proposed SL-Diff yields excellent prediction results within a reasonable sampling time, as demonstrated in extensive experiments on five datasets.
\end{abstract}

\section{Introduction}

The exponential growth of network-structured information has aroused widespread interest in studying its dissemination mode \cite{Inf-VAE2020,deepis,H-Diffu2022,Dydiff-vae2021}. This involves modeling the  information dissemination (information diffusion\footnote{In most studies, this problem is referred to as \enquote{graph information diffusion}. However, we refer to it as \enquote{graph information dissemination} in this paper to disambiguate with the \enquote{diffusion model}.}) based on the corresponding network structure and dissemination source.
While there are few research followers, the inverse problem of information dissemination on graphs and source localization is of great practical significance. For instance, locating the source account for spreading rumors in social networks is crucial for rumor detection \cite{LPSI,GCNSI}. Similarly, source localization can aid in virus interception \cite{IVGD}, malicious email traceability \cite{Rapid2022}, and other areas \cite{vaeSL2022,OJC}.

\begin{figure}[t]
      \begin{minipage}[c]{0.79\textwidth}
        \centering
        \vspace{0.3cm}
        \includegraphics[width=\linewidth]{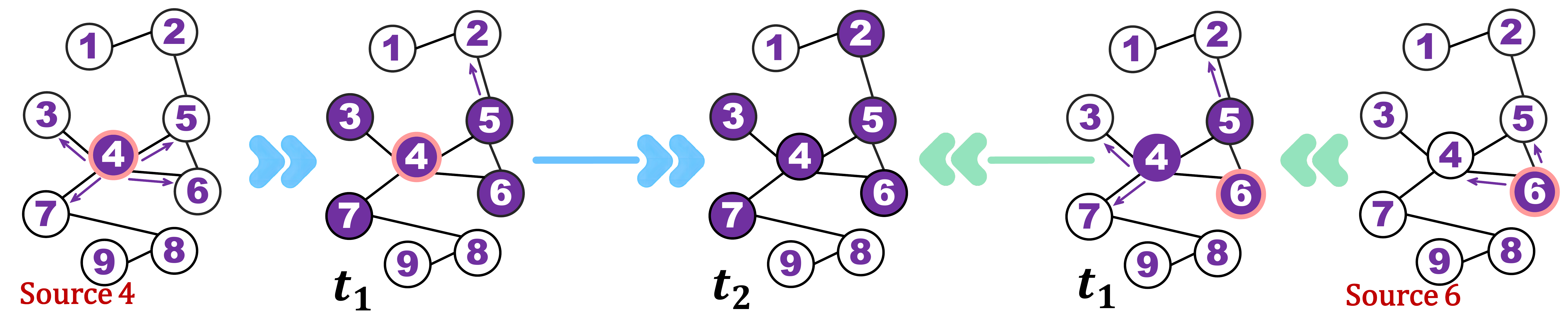}
        \caption{Many-to-one (ill-posed) relationship when the supervised signal is the source indicator.}
        \label{fig:illposed1}
        \end{minipage}
        \vspace{1.5cm}
        \centering
        \begin{minipage}[c]{0.87\textwidth}
          \centering
          \includegraphics[width=\linewidth]{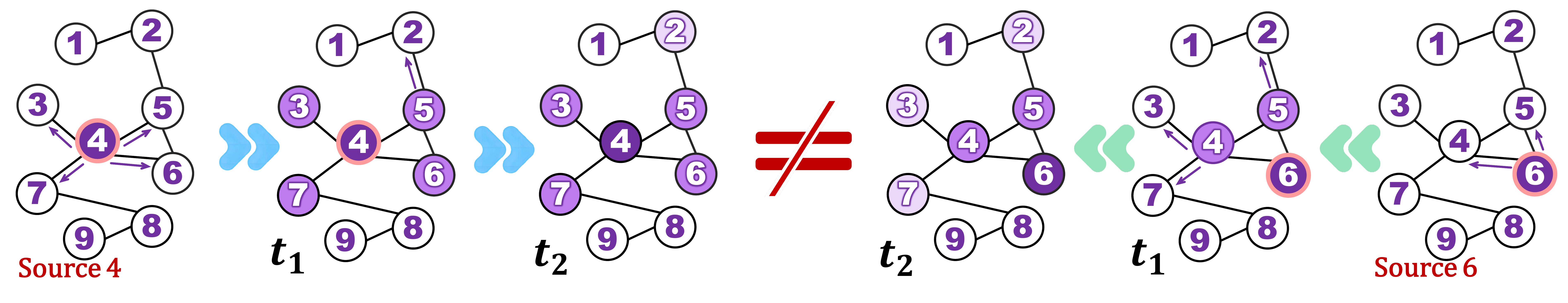}
          \caption{One-to-one relationship when the supervised signal is the source proximity degree.}
          \label{fig:illposed2}
          \end{minipage}
  \end{figure}
  \begin{wrapfigure}{r}{5cm}
    \centering
   \includegraphics[width=0.7\linewidth]{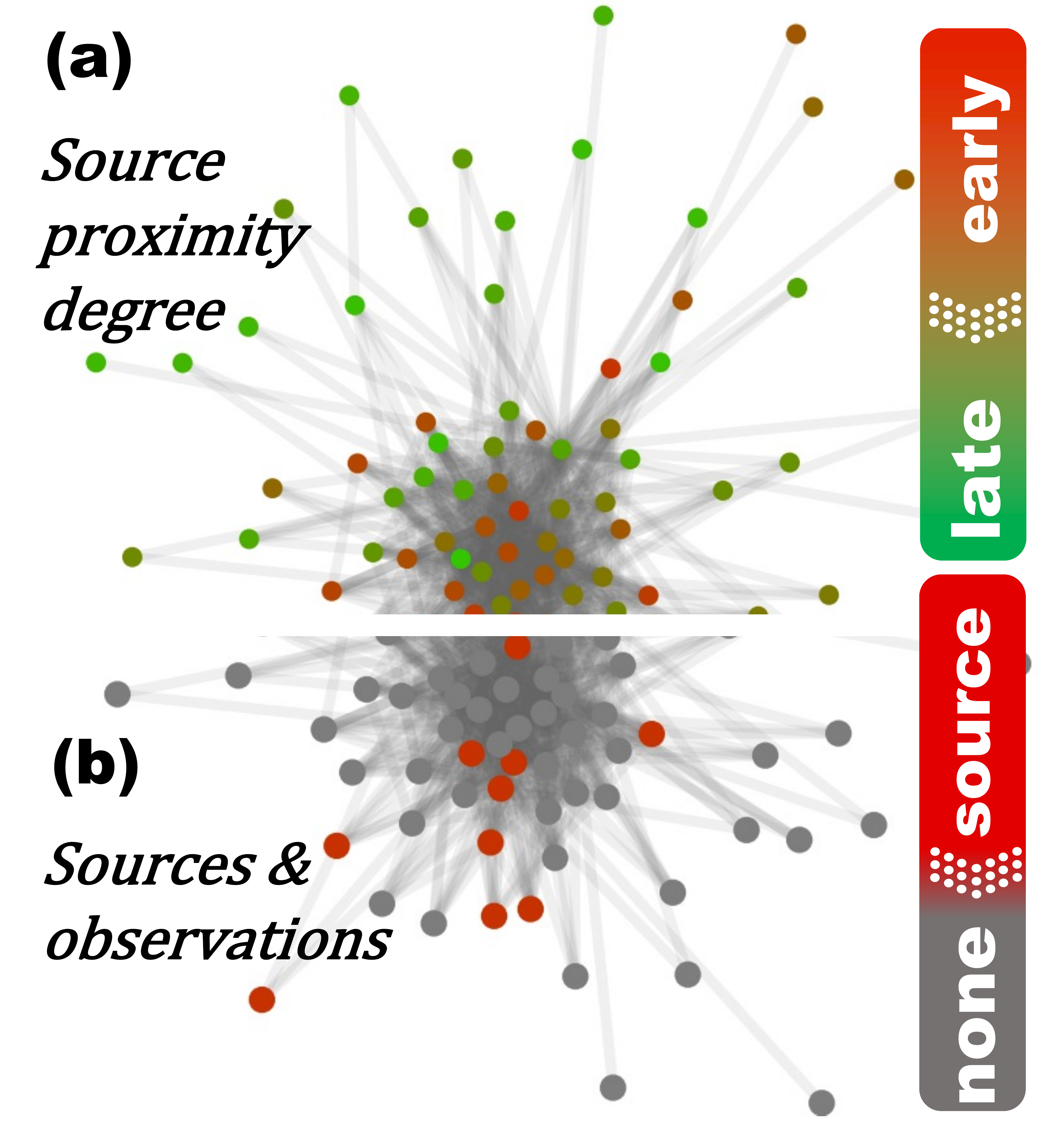}
         \vspace{-0.5cm}
         \caption{ Two types of supervised signal derived from a cascade.}
         \label{fig:graph}
  \end{wrapfigure}
 However, there are currently several challenges hindering this field: \textbf{Chall. 1.} Existing models do not fundamentally address the ill-posed problem because their training data are inherently ill-posed (many-to-one). For instance, in the case of blurred image restoration, the ill-posed problem cannot be solved if the training data consist solely of blurred images.   In fact, existing methods unfortunately erase the relative temporal correlation of each node from the cascades when setting the training data, which essentially leads to the ill-posed problem of source localization. This abandonment of temporal information also significantly restricts the ability of the model to learn the underlying dissemination pattern. \textbf{Chall. 2.} In practical applications, information dissemination in the network will introduce many uncertain factors, that is, the noise irrelevant to the regular dissemination mode. Although some studies \cite{vaeSL2022,Dydiff-vae2021} have begun to discuss the uncertainty in source localization, how to quantitatively model them and more importantly, eliminate the relevant noise in source localization is unresolved.

The denoising diffusion models \cite{ddpm,Score-BasedSong,posteriorsampling} have flourished recently due to their high-quality reconstructions and powerful inverse problem solving abilities. In addition to leveraging its power and further addressing Chall. 2, we have taken the first step in generalizing this powerful model to the source localization problem. However, the existing family of diffusion models poses the following challenges for direct migration. \textbf{Chall. 3.} As the inverse problem solver, existing diffusion models need to calculate the conditional posterior probability at each diffusion step for Maximum A Posteriori (MAP) approximation. However, in the case of source localization, the calculation of posterior probability involves simulating information dissemination across the entire network structure at each diffusion step, making it extremely computationally expensive. Therefore, it is almost impossible to directly apply existing diffusion models to source localization. \textbf{Chall. 4.} Existing diffusion models \cite{ddpm,Score-BasedSong,sliced2020song,GeoDiff} that were designed for image or molecule data struggle to model dissemination patterns. For example, the number of nodes in the molecular graph is much smaller than that in  the information dissemination network, and does not have the temporal information.

\begin{figure}[t]
  \centering
  \begin{minipage}[c]{1\textwidth}
  \centering
  \includegraphics[width=\linewidth]{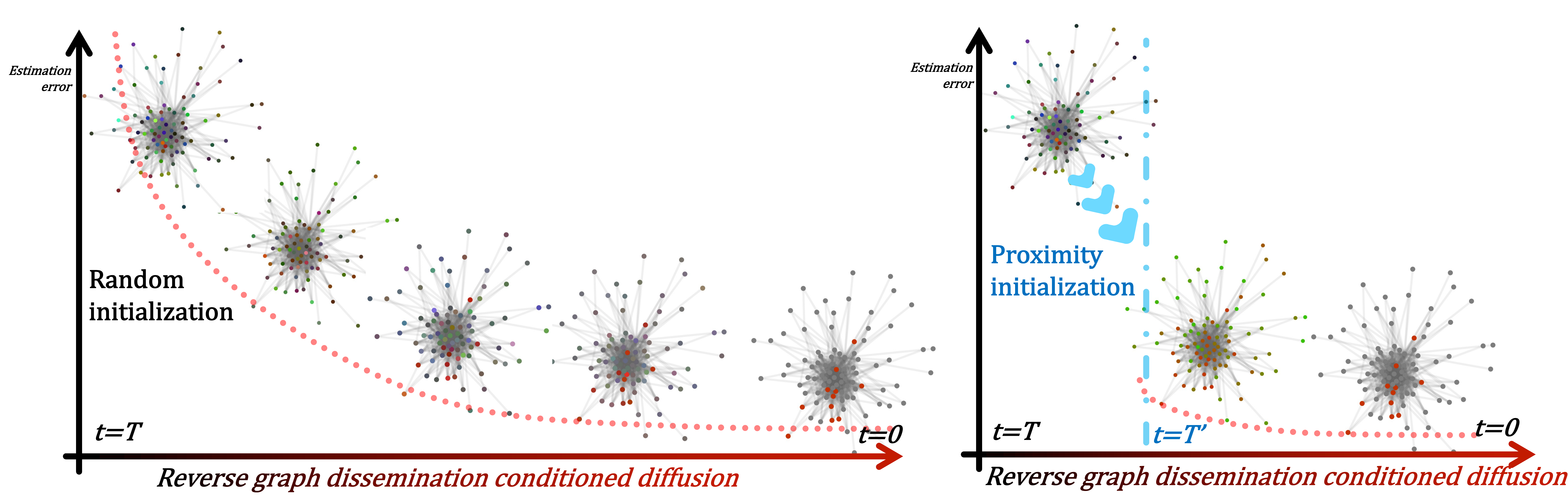}
  \caption{Random initialization and proximity initialization of sampling.}
  \label{fig:error}
  \end{minipage}
\end{figure}
To address the complex challenges outlined above and effectively harness the benefits of current technological developments, we propose a two-stage optimization framework (Figure \ref{fig:two}) named Source Localization Denoising Diffusion model (SL-Diff).
In general, SL-Diff leverages an efficient initialization that is supervised with temporal cascade information and an exquisitely redesigned denoising network structure to achieve a series of outstanding performances.
Specifically, for Chall. 1, we ingeniously retain the timing of the cascade (defined as the source proximity degree, elaborated in Section \ref{sec:GraphInformationdissemination}) during the model training phase while remaining consistent with baselines that only take the disseminated observation as the input for prediction during testing. This means that the model in the coarse stage aims to learn the source proximity relationship (i.e., a kind of one-to one mapping correlation) through training, which directly addresses the core of Chall. 1.
On the other hand, due to the contraction property of stochastic differential equations (SDEs), the error reduction of the diffusion model is exponential. Based on this characteristic, we propose a two-stage optimization framework. In the the coarse stage, we use the one-shot method to generate coarse source proximity degrees quickly and efficiently, reducing the predicted source localization error. In the  the fine stage, the diffusion model simulates information dissemination across the entire network, accurately locating the source. Concretely, we optimize the verbose sampling process (Figure \ref{fig:error} (left)) into an efficient two-stage process (Figure \ref{fig:error} (right)) and further balance the two-stage diffusion step ratio to achieve an optimum between efficiency and accuracy through parameter experiments. This approach effectively solves Chall. 3.
Since the coarse stage provides excellent initialization of the coarse source proximity relationships, it prominently alleviates the local optimum problem caused by class imbalance in the source localization problem.
In addition, we propose a new uncertainty graph information dissemination model that quantizes the dissemination noise and a score function approximating network that adapts to the underlying dissemination mode on the graph to jointly address Chall. 2 and Chall. 4.
In summary, our contributions are as follows:

\renewcommand{\labelitemi}{\textbullet}
\begin{itemize}
  \item We propose a training framework that employs source proximity degree supervision, which fundamentally addresses the ill-posed problem of source localization.

  \item Our contribution lies in proposing, for the first time, a two-stage denoising diffusion model for the source localization problem. The targeted design of the two stages effectively addresses the challenge of migrating the diffusion model to this field.
  
  \item  We design a new model that approximates graph information dissemination and a score function approximating network to enhance the performance of the diffusion model for source localization.
  
  \item Experiments on five real-world datasets demonstrate that our proposed SL-Diff model outperforms state-of-the-art models.  The code for this study is available  at  \url{https://anonymous.4open.science/r/SL-Diff-275A}.
\end{itemize}


\section{Related Work}
\subsubsection{Graph Information Dissemination and Source Localization.}

Graph information dissemination modeling aims to predict the nodes to be affected given the source nodes, which is one of the foremost technologies in social network analysis, disease infection prediction, etc. Traditional methods \cite{Maximizingthespread2003,Networksandepidemic2005,Apeek2013} manually model the respective dissemination pattern for different application fields while suffering from poor generalizability and high computational complexity. With the blooming of deep learning, \cite{CascadeDynamicsModelin2017,Amultimodalvariational2020,Dydiff-vae2021} incorporate recurrent neural networks to capture the dynamic relationship of dissemination cascades. Graph neural networks have further been introduced to aggregate the node neighboring information and model the dissemination pattern to facilitate prediction.

Source localization aims to infer source nodes given an observed set of nodes, which has essential applications in rumor tracing and infection source discovery. Generally speaking, source localization is used to study the traceability method based on the dissemination model. Early studies \cite{sourcedetectionSIR2014,OJC,GCNSI} were based on specific dissemination models such as Susceptible-Infected (SI) and  Susceptible-Infected-Recovered (SIR). \cite{LPSI} argues that a fixed propagation model to be preset is not necessary. Furthermore, \cite{Rapid2022} focuses on detecting the source in the early propagation stage to reduce the loss caused by propagation. \cite{IVGD} develops a framework for the inverse of graph dissemination models to detect the source. From another perspective, \cite{vaeSL2022} introduces VAE \cite{VAE2013} to probabilistically model the uncertainty in source localization.
\subsubsection{Diffusion Models.}
The diffusion model we discuss in this paper is the score-based generative model \cite{Score-BasedSong}, which applies the stochastic differential equation (SDE) to learn the gradient of the target distribution. The currently popular DDPM \cite{ddpm} is a particular case of it. As the diffusion model separates the noise from the data step by step at a fine-grained level, its powerful generative capabilities have achieved state-of-the-art results in many fields, e.g.,  image generation \cite{sliced2020song,ddpm,Score-BasedSong,posteriorsampling}, graph generation \cite{SDEGraphs2022,GeoDiff}, and time series generation \cite{Autoregressivedenoising2021}.

\section{Preliminaries}

\subsection{Conditional Score-Based Diffusion Models}

Diffusion models aim to approximate the prior distributions by learning the noise of the data reversely.  \cite{Score-BasedSong} combines SMLD \cite{sliced2020song} and DDPM \cite{ddpm} into a generalized theoretical framework, known as the Score-Based Diffusion Model. They generally map data to a noise distribution (the prior) with a  stochastic differential equation (SDE), and reverse this SDE for generative modeling. 
\subsubsection{Forward SDE.}
Given the i.i.d. original dataset samples $\mathbf{x}(0) \sim p_\mathbf{x}$,  which are further indexed as $\mathbf{x}(t)^T_{t=0}$ by the diffusion step $t \in [0,T]$ to indicate the noise degree, the diffusion process can be modeled as the solution to an It\^{o} SDE:
\begin{equation}
d \mathbf{x}=-\frac{\beta(t)}{2} \mathbf{x} d t+\sqrt{\beta(t)} d \mathbf{w}
\end{equation}
where $\beta(t)\in \mathbb{R} $ is the noise schedule that we uniformly adopt the one in \cite{ddpm} in this paper, $\mathbf{w}$ is the Brownian motion.
\subsubsection{Reverse SDE.}
\cite{anderson1982reverse} clarifies that the reverse of a diffusion process is also a
diffusion process, which can be modeled as the reverse SDE:

\begin{equation}\label{equ:ReverseSDE}
  d \mathbf{x}=\left[-\frac{\beta(t)}{2} \mathbf{x}-\beta(t)\nabla_{\mathbf{x}_t} \log p_t\left(\mathbf{x}_t\right)\right] d t+\sqrt{\beta(t)} d \bar{\mathbf{w}},
\end{equation}
where  $dt$ is the negative diffusion step from $T$ to $0$ and $\bar{\mathbf{w}}$ is the corresponding Brownian motion of the reverse process. Since the direct estimation of $ \nabla_{\mathbf{x}_t} \log p_t\left(\mathbf{x}_t\right)$ is too computationally intensive and badly generalizatial, we train a  score-based model to approximate it:
\begin{equation}
\begin{aligned}
\theta^*&=\underset{\theta}{\arg \min } \mathbb{E}_{t \sim U(\varepsilon, 1), \mathbf{x}(t) \sim p(\mathbf{x}(t) \mid \mathbf{x}(0)), \mathbf{x}(0) \sim p_{\text {data }}}[\xi]
\\
\xi&=\left\|\mathbf{s}_\theta(\mathbf{x}(t), t)-\nabla_{\mathbf{x}_t} \log p(\mathbf{x}(t) \mid \mathbf{x}(0))\right\|_2^2
\end{aligned}
\end{equation}
Here $\varepsilon \simeq 0$ is the small positive constant. 
After acquiring the trained score function approximation network, the initial noise can be denoised to enable sampling based on specific posterior conditions, leading to the generation of expected data.

\subsection{Graph Information dissemination}\label{sec:GraphInformationdissemination}

The graph information dissemination problem involves a graph $G=(V,E)$ with edge set $E$ and node set $V$. An information dissemination cascade $D_i$ at length $K+1$ on the graph is defined as $D_i=\left\{\left(v_{i_k}, k\right) \mid v_{i_k} \in V, k=0,1 \ldots ,K-1, K\right\}$, where $k$ is non-decreasing and the first $K_s$ nodes are defined as source nodes. The source indicator $\mathbf {i} \in \{ 0,1 \}^{|V|}$ is defined as $0$ for being the source and $1$ for not, while the source proximity degree $\mathbf{x} \in [0,1]^{|V|}$ is defined as $x_k=\frac{k}{K}$. The set of affected nodes' observations is denoted as $\mathbf{y} \in \{ 0,1 \}^{|V|}$, where $0$ indicates being affected and $1$ indicates not being affected. The goal of the graph information dissemination problem is to predict the affected nodes given the source indicator and graph structure.

Thanks to the outstanding performance of GNN on graph data, the state-of-the-art graph information dissemination methods \cite{Inf-VAE2020,Dydiff-vae2021,H-Diffu2022,MS-HGAT2022} construct various realistic-meaning attribute variables through GNN at the first stage, and then perform variable dissemination to derive the final affected node sets. Specifically: (1) In the variable construction, we define a neural network $ \mathbf{v}=g_{\mathbf{w_1}}(\mathbf{x})$ to construct miscellaneous variables of nodes (such as sender variable or receiver variable). (2) In variable dissemination, we define a dissemination neural network $ f_{\mathbf{w_2}}(\mathbf{v})$ to propagate information to neighbour nodes according to the topology of the graph, where $w_1$ and $w_2$ are learnable parameters. Thus, the general paradigm of graph information dissemination can be defined as:
\begin{equation}\label{equ:desseminate}
\mathbf{y}=f_{\mathbf{w_2}}\left(g_{\mathbf{w_1}}(\mathbf{x})\right)
\end{equation}
\subsection{Problem Formulation for Source Localization}
In order to eliminate ambiguity, the graph mentioned in Subsection \ref{sec:GraphInformationdissemination} is referred to as the whole graph $\mathcal{G}^w=(V^w,E^w)$.
Source localization is the corresponding inverse problem of graph information dissemination, which is formally defined
as: given the whole graph information $\mathcal{G}^w=(V^w,E^w)$ and  affected nodes $\mathbf {y}  \in \{ 0,1 \}^{|V|}$, reconstruct the source nodes $\mathbf {\hat i} \in \{ 0,1 \}^{|V|} $.
For the convenience of mentioning later, we define the cascade graph as $\mathcal{G}^c_i=(V^c_i\in D_i,E^c )$ where $E^c=\{(u, v) \in E^w \mid u \in D_i \text{ or } v \in D_i \} $.

\section{SL-Diff Method}

\subsection{Two-stage Optimization Framework}
\begin{wrapfigure}{r}{5cm}
  \centering
\includegraphics[width=\linewidth]{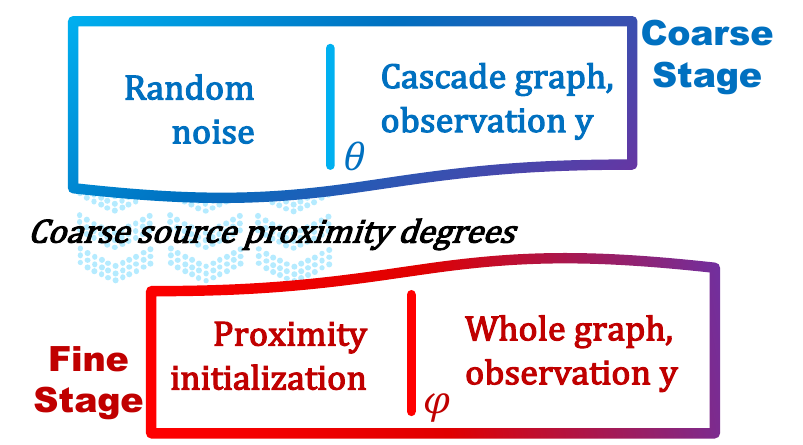}
  \caption{Two-stage optimization framework}
  \label{fig:two}
\end{wrapfigure}
To mitigate the computational burden of simulating information dissemination on the whole graph at each diffusion step (chall. 3), and to leverage the cascade temporal information for addressing the ill-posed problem (chall. 3), we propose a two-stage optimization framework, as depicted in Figure \ref{fig:two}. Specifically, in the coarse stage, the supervising signals for the model are the source proximity degrees (Figure \ref{fig:graph}(a)) that fully retain the cascade temporal information. In other words, the final outputs of this stage are the relatively coarse-grained node infection sequence predictions, which are used to initialize the precise source localization for the next stage efficiently. In the second stage, i.e., the fine stage, the supervising signals of model training are source indicators (Figure \ref{fig:graph}(b)), which aim to accurately output the predicted source under the given disseminated observation conditions. In this stage, the model simulates the information dissemination on the graph at each diffusion step to calibrate predictions.

\subsection{Coarse Proximity Generation}

\begin{figure}[t] 
  \centering
  \includegraphics[width=0.9\linewidth]{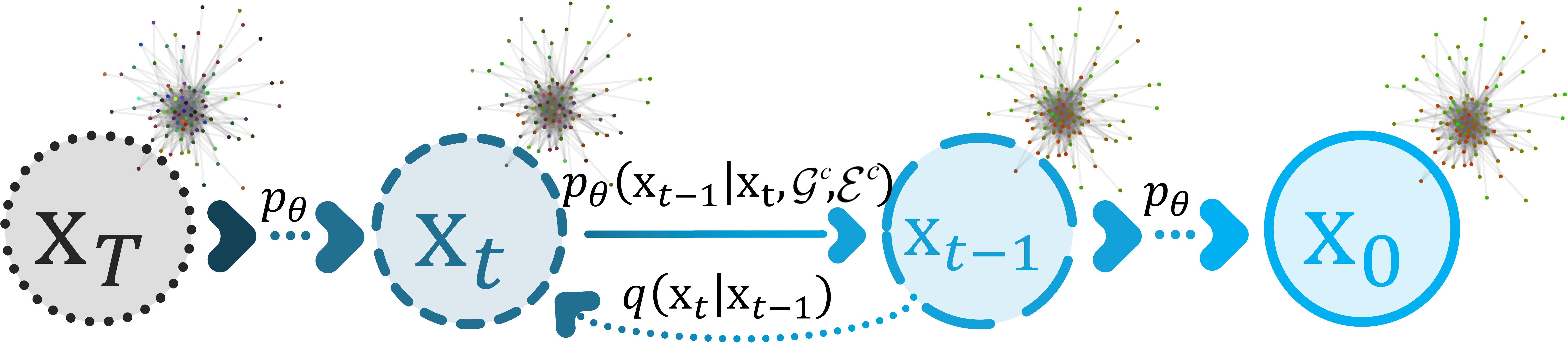}
  \caption{The forward diffusion and reverse process of the coarse stage.}
  \label{fig:initialization}
\end{figure}
In this section, we propose a coarse proximity generation model at the first stage, which serves to effectively initialize the subsequent stage. The \enquote{coarse} is used in two senses. Firstly, the source proximity degree derived from the cascade data does not strictly reflect the infection relationships between nodes, as the nodes in a cascade may be infected by their common ancestors rather than by their directly adjacent forward nodes. Secondly, as no dissemination model is incorporated at this stage, it is not feasible to accurately predict the dissemination source in reverse. Nonetheless, the generated coarse source proximity degrees are well-suited for the initialization of the downstream diffusion model. Although the coarse source proximity generation occurs only within the cascade graph, it is incomplete without considering the structure of the entire graph. This is due to the significant impact that the relative position of each node in the graph has on information dissemination. To address this, we utilize Position-aware Graph Neural Networks (P-GNN) \cite{pgnn} to perform node position representation learning on the whole graph of each dataset. Through this process, we obtain the positional embeddings $\mathcal{E}^w=\{\mathbf{e}_i\mid \mathbf{e}_i=\text{P-GNN}(v_i), v_i \in V^w\} $ that reflect the relative position of each node with respect to other nodes on the whole graph. We use this positional representation in the specific cascade $\mathcal{E}^c=\{\mathbf{e}_i\mid \mathbf{e}_i=\text{P-GNN}(v_i), v_i \in V^c\} $ as a conditional input to the score function approximation network, enhancing its ability to fit the underlying features of the cascades.

Let the original source proximity degrees derived from a cascade   be denoted as $\mathbf{x}_0$. The reverse SDE here follows the Equation \ref{equ:ReverseSDE}, while the score function  conditioned on the the cascade graph structure and the positional embeddings  is changed to $\mathbf{s}_{\boldsymbol{\theta}}(\mathbf{x}_t, t \mid \mathcal{G}^c, \mathcal{E}^c)$. We adopt the same variance schedule as DDPM \cite{ddpm} to discretize the formula,  then the estimated reverse Markov chain is defined as follows:

\begin{equation}
  \label{equ:xi}
  \mathbf{x}_{t-1}=\frac{1}{\sqrt{1-\beta_t}}\left(\mathbf{x}_t+\beta_t \mathbf{s}_{\boldsymbol{\theta}^*}\left(\mathbf{x}_t, t \mid \mathcal{G}^c ,\mathcal{E}^c\right)\right)+\sqrt{\beta_t} \mathbf{z}_t, \quad t=T, T-1, \cdots, 1
  \end{equation}
  in which we use the following re-weighted variant of the evidence lower bound (ELBO) to train the model to obtain the estimated score function:

\begin{equation}
  \boldsymbol{\theta}^*=\underset{\boldsymbol{\theta}}{\arg \min } \sum_{t=1}^N\left(1-\alpha_t\right) \mathbb{E}_{p_{\text {data }}(\mathbf{x})} \mathbb{E}_{p_{\alpha_t}({\mathbf{x}_t} \mid \mathbf{x})}\left[\left\|\mathbf{s}_{\boldsymbol{\theta}}(\mathbf{x}_t, t \mid \mathcal{G}^c, \mathcal{E}^c)-\nabla_{\mathbf{x}_t} \log p_{\alpha_t}(\mathbf{x}_t \mid \mathbf{x})\right\|_2^2\right]
  \end{equation}
 where $\alpha := \prod^{i}_{j=1}(1- \beta_j)$. 
 
 To efficiently approximate the  score functions, we propose a new network structure that properly introduces conditional position embeddings while preserving nodes' adjacent characteristics at different orders.  Our approach utilizes a multi-head graph attention (GMT) \cite{gmt} as the basic operation of graph convolution, allowing us to aggregate information across $L$-layer graphs. Furthermore, we leverage skip connections between each layer to improve the flow of information. Finally, we concatenate the output of each layer and perform multi-layer nonlinear transformations to obtain an estimated score function. The proposed conditional score function approximating network is defined as follows:

\begin{equation}
  \label{equ:approximation}
  \begin{aligned}
  \mathbf{H}^{l+1}&=\text{GMT}(\mathbf{H}^l,\mathcal{G}^c)+\mathbf{H}^l 
  \\
  \mathbf{s}_{\boldsymbol{\theta}}(\mathbf{x}_t, t \mid \mathcal{G}^c, \mathcal{E}^c)&=\text{MLP}\left(\mathbf{catenate}[\mathcal{E}^c,\mathbf{H}^0,\dots,\mathbf{H}^{L-1,}, \mathbf{H}^L,\mathcal{T}(t)]\right) 
\end{aligned}
\end{equation}
where  $\mathbf{H}^0 =\mathbf{x}_t+\mathcal{E}^c$. We use the positional encodings \cite{vaswani2017attention}  to encode the diffusion step, and the formula is defined as: $\mathcal{T}(t)=\left[\ldots, \cos \left(t / r^{\frac{-2 d}{D}}\right), \sin \left(t / r^{\frac{-2 d}{D}}\right), \ldots\right]^{\mathrm{T}}$, where $d=1,\dots, D/2$ is the  dimension of the
embedding, and r is a large constant (set to $10^5$). After obtaining the trained network $\mathbf{s}_{\boldsymbol{\theta}}$, we sample according to the Algorithm \ref{al:sampling1} and get the predicted coarse source proximity degrees.

\subsection{Graph Dissemination Conditioned Model}
\begin{figure}[t]
  \centering
  \hspace{-1cm}\begin{minipage}[c]{0.65\textwidth}
  \centering
  \includegraphics[width=\linewidth]{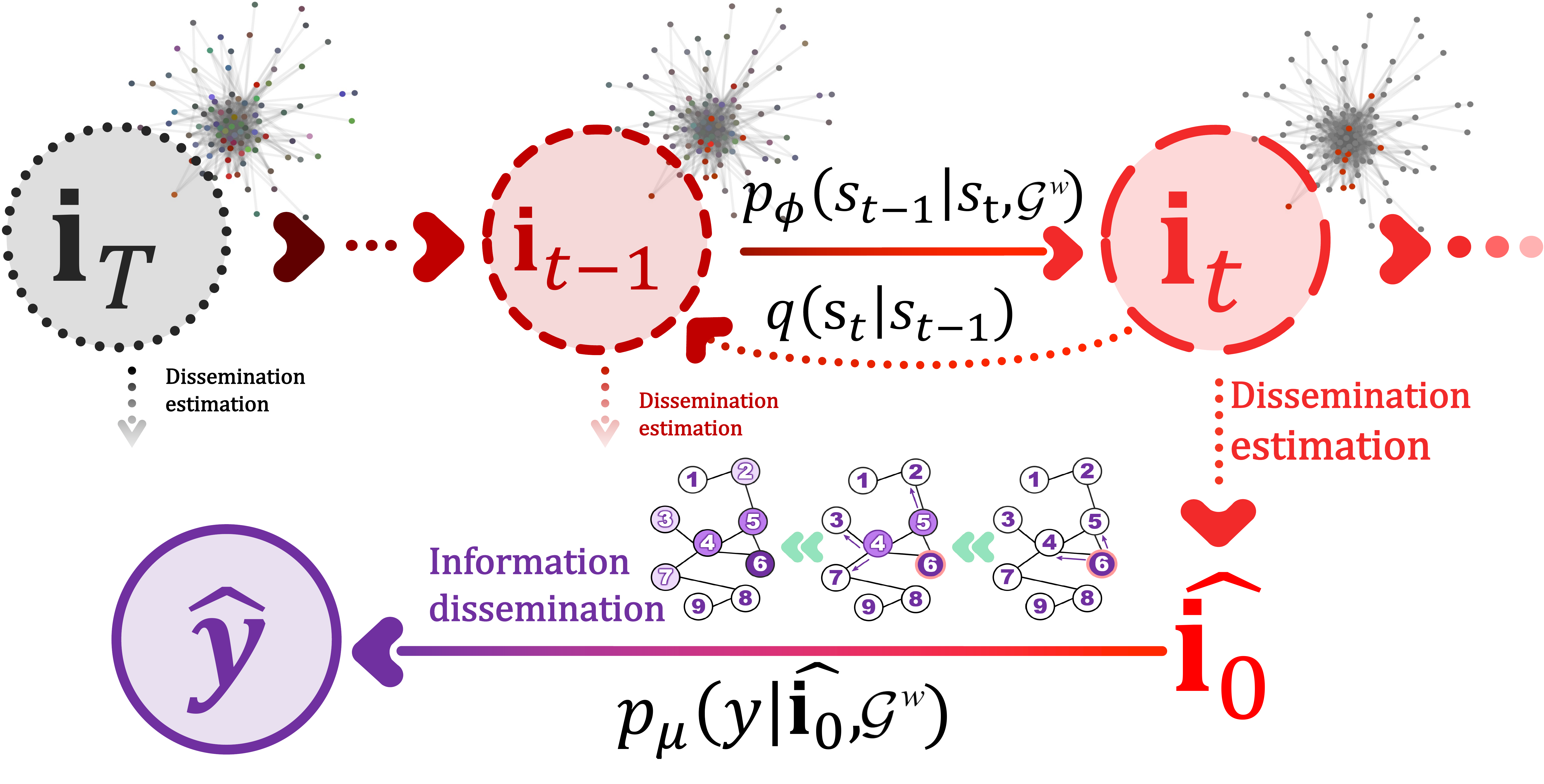}
  \caption{The forward diffusion and reverse process of the fine stage.}
  \label{fig:stage2}
  \end{minipage}
\end{figure}
In the second stage, we apply the estimated source proximity degree to initialize the graph dissemination conditioned diffusion model. Significantly different from the conditional diffusion in the first stage, the diffusion model here is an inverse problem solver, which aims to recover the source indicator $\mathbf{i}$ from the disseminated observation $\mathbf{y}$. The connection between them is the forward information dissemination process of $\mathbf{i} \rightarrow \mathbf{y}$.

If the assumption is the same as in \cite{Score-BasedSong}, that is, $p_t\left(\mathbf{y} \mid \mathbf{i}_t \right)$ is tractable, then the reverse SDE of the inverse problem is defined as:

\begin{equation}
  \label{equ:di}
  d \mathbf{i}=\left[-\frac{\beta(t)}{2} \mathbf{i}-\beta(t)\left(\nabla_{\mathbf{i}_t} \log p_t\left(\mathbf{i}_t\right)+\nabla_{x_t} \log p_t\left(\mathbf{y} \mid \mathbf{i}_t , \mathcal{G}^w\right)\right)\right] d t+\sqrt{\beta(t)} d \bar{\mathbf{w}},
\end{equation}
However, in our case, $p_t\left(\mathbf{y} \mid \mathbf{i}_t \right)$ is intractable, as the dissemination model cannot support the source with noise as input. To bridge this gap, we introduce the approximation method \cite{posteriorsampling} for $ p(\hat{\mathbf{i}_0 }\mid\mathbf{i}_t)$:

\begin{equation}\label{equ:estmateI}
  \hat{\mathbf{i}_0} \simeq \frac{1}{\sqrt{\bar{\alpha}(t)}}\left(\mathbf{i}_t+(1-\bar{\alpha}(t)) \boldsymbol{s_{\phi}}\left(\mathbf{i}_t, t \mid \mathcal{G}^w\right)\right) 
  \end{equation}
where $\alpha_i \triangleq 1-\beta_i, \bar{\alpha}_i \triangleq \prod_{j=1}^i \alpha_i$ following \cite{ddpm}. 
Now the problem lies in determining an appropriate dissemination model to calculate the posterior probability of $\mathbf{y}$ from estimated $\mathbf{i}$. In real-world applications, various random factors can impact the information dissemination process, which are not accounted for in the basic dissemination function (Equation \ref{equ:desseminate}). Our proposed model addresses this challenge by qualifying the uncertain trough Gaussian noise $\tilde{\sigma}(t)$ with adaptive variance (parameterized by MLP and taking diffusion step as input). Additionally, our model can eliminate interference from noise to source localization through the powerful iterative denoising process of the diffusion model. Thus, the formula is defined as follows:

\begin{equation}\label{equ:noise}
  \mathbf{y}=f_{\mathbf{}}\left(g_{\mathbf{}}(\mathbf{i})\right) + \tilde{\sigma}(t)\boldsymbol{\epsilon},\qquad \boldsymbol{\epsilon} \sim  \mathcal{N}(\mathbf{0,I})
  \end{equation}
From the above formula, we can calculate the partial derivative $\nabla_{\mathbf{i}_0} \log p\left(\mathbf{y} \mid \mathbf{i}_0\right)$. And bring the conclusion in Equation 8 into it to get an approximate  for the derivative of $p\left(\mathbf{y} \mid \mathbf{i}_t\right)$ with respect to $\hat{\mathbf{i}_t}$.

  \begin{equation}
    \nabla_{\hat{\mathbf{i}_t}} \log p\left(\mathbf{y} \mid \mathbf{i}_t\right) \simeq-\frac{1}{{\tilde{\sigma}(t)}^2} \nabla_{\mathbf{i}_t}\left\|\mathbf{y}-\hat{\mathbf{y}}\left(\hat{\mathbf{i}}_0\left(\mathbf{i}_t\right)\right)\right\|_2^2
    \end{equation}
Hence, we can discretize Equation \ref{equ:di} similarly to formula \ref{equ:xi} and deduce the expression for $p(\mathbf{i}_{t-1}) \mid \mathbf{i}_t)$. This allows us to perform iterative denoising of $\mathbf{i}$ given the posterior $\mathbf{y}$, ultimately leading to an accurate estimation of $\mathbf{i}$. For a more thorough explanation of this procedure, please refer to  Algorithm \ref{al:sampling2}.
\begin{minipage}[ht]{0.5\textwidth}
  \hspace{-0.1cm}
  \begin{algorithm}[H]
    \caption{Sampling of the coarse stage}
    
    \begin{algorithmic}[1]\label{al:sampling1}
\REQUIRE 
ground truth source proximity degree ${\mathbf x}$,
number of diffusion steps $T_1$,
variance schedule $\beta_t$,
the cascade graph $ \mathcal{G}_c$
 \STATE $\mathbf{x}_{T_1} \sim \mathcal{N}(\mathbf{0,I})$
\FOR{$t=T_1-1$ \TO $t=0$ } 

  \STATE $ \hat{\mathbf{s}} \leftarrow\mathbf{s}_{\boldsymbol{\phi}}(\mathbf{x}_t, t \mid \mathcal{G}^w, \mathcal{E}^w)$
  \STATE $\mathbf{x}_{t-1}' \leftarrow (2-\sqrt{1-\beta
  _t}\mathbf{x}_{t}) + \beta
  _t\left(\hat{\mathbf{s}}\right) $
  \STATE $\mathbf{z} \sim \mathcal{N}(\mathbf{0,I})$
  \STATE $\mathbf{x}_{t} \leftarrow \mathbf{x}_{t-1}' + \sqrt{\beta_t} \mathbf{z} $

  \ENDFOR
\RETURN the estimated $\mathbf{x}_{0}$
\end{algorithmic}
  \end{algorithm}
  \end{minipage}
  \hspace{-0.cm}
  \begin{minipage}[ht]{0.5\textwidth}
  \centering
    \begin{algorithm}[H]
        \caption{Sampling of the fine stage}
        
        \begin{algorithmic}[1]\label{al:sampling2}
    \REQUIRE disseminated observation $\mathbf{y}$,
    coarse source proximity degree $\hat{\mathbf x}$,
   number of diffusion steps $T_2$,
   variance schedule $\beta_t$,
   the whole graph $ \mathcal{G}_W$

   \STATE $\mathbf{i}_{T_2} \leftarrow \hat{\mathbf x}$
   \FOR{$t=T_2-1$ \TO $t=0$ } 
      \STATE $ \hat{\mathbf{s}} \leftarrow\mathbf{s}_{\boldsymbol{\phi}}(\mathbf{i}_t, t \mid \mathcal{G}^w)$
  
      \STATE $\mathbf{i}_{t-1}' \leftarrow (2-\sqrt{1-\beta
      _t}\mathbf{i}_{t}) + \beta
      _t\left(\hat{\mathbf{s}}-\frac{1}{{\tilde{\sigma}(t)}^2} \nabla_{\mathbf{i}_t}\left\|\mathbf{y}-\hat{\mathbf{y}}\left(\hat{\mathbf{i}}_0\left(\mathbf{i}_t\right)\right)\right\|_2^2 \right) $
      \STATE $\mathbf{z} \sim \mathcal{N}(\mathbf{0,I})$
      \STATE $\mathbf{i}_{t} \leftarrow \mathbf{i}_{t-1}' + \sqrt{\beta_t} \mathbf{z} $

      \ENDFOR
    \RETURN the estimated $\mathbf{i}_{0}$
  \end{algorithmic}
      \end{algorithm}
  \end{minipage}

  \begin{table}[t]
    \caption{Model performance across five datasets.}
    \label{tb:all}
   \centering\scriptsize
  \begin{tabular*}{\textwidth}{@{\extracolsep{\fill}}c|c|c|c|c|c|c|c|c}
    \midrule[1.5pt]
            Datasets              & Methods   & Netsleuth & OJC    & LPSI   & GCNSI  & IVGD              & SL-VAE & SL-Diff           \\  \midrule[0.3pt]
    \multirow{4}{*}{Digg}         & RE     & 0.0142    & 0.0781 & 0.2352 & 0.0135 & 0.2310            & 0.5420 & $\mathbf{0.7813}$ \\   
                                  & PR     & 0.0023    & 0.0554 & 0.0072 & 0.2369 & 0.1397            & 0.4216 & $\mathbf{0.5839}$ \\   
                                  & F1     & 0.0040    & 0.0648 & 0.0140 & 0.0255 & 0.1741            & 0.4743 & $\mathbf{0.6683}$ \\   
                                  & ACC    & 0.7714    & 0.9035 & 0.9531 & 0.8064 & 0.9327            & 0.9742 & $\mathbf{0.9824}$ \\  \midrule[0.3pt]
    \multirow{4}{*}{Memetracker}  & RE     & 0.0647    & 0.0256 & 0.3047 & 0.2953 & 0.5954            & 0.5010 & $\mathbf{0.6902}$ \\   
                                  & PR     & 0.0247    & 0.0360 & 0.1145 & 0.0172 & 0.1556            & 0.4592 & $\mathbf{0.4721}$ \\   
                                  & F1     & 0.0358    & 0.0299 & 0.1665 & 0.0325 & 0.2467            & 0.4792 & $\mathbf{0.5607}$ \\   
                                  & ACC    & 0.5688    & 0.6675 & 0.9174 & 0.8428 & 0.8947            & 0.9420 & $\mathbf{0.9562}$ \\  \midrule[0.3pt]
    \multirow{4}{*}{Android}      & RE     & 0.3172    & 0.1401 & 0.3407 & 0.7434 & 0.7253            & 0.6261 & $\mathbf{0.8260}$ \\   
                                  & PR     & 0.0422    & 0.0610 & 0.2323 & 0.3024 & 0.4105            & 0.5284 & $\mathbf{0.5945}$ \\   
                                  & F1     & 0.0745    & 0.0850 & 0.2762 & 0.4299 & 0.5243            & 0.5731 & $\mathbf{0.6914}$ \\   
                                  & ACC    & 0.6215    & 0.8337 & 0.9404 & 0.8211 & 0.9530            & 0.9245 & $\mathbf{0.9937}$ \\  \midrule[0.3pt]
    \multirow{4}{*}{Christianity} & RE     & 0.2491    & 0.3478 & 0.5309 & 0.7294 & 0.6433            & 0.8011 & $\mathbf{0.8352}$ \\   
                                  & PR     & 0.1184    & 0.2823 & 0.6249 & 0.2300 & $\mathbf{0.5202}$ & 0.4894 & 0.5120            \\   
                                  & F1     & 0.1605    & 0.3116 & 0.5741 & 0.3497 & 0.5752            & 0.6076 & $\mathbf{0.6348}$ \\   
                                  & ACC    & 0.7140    & 0.9304 & 0.9122 & 0.9673 & 0.9781            & 0.9529 & $\mathbf{0.9818}$ \\  \midrule[0.3pt]
    \multirow{4}{*}{Twitter}      & RE     & 0.0184    & 0.0154 & 0.2091 & 0.3770 & 0.6219            & 0.3273 & $\mathbf{0.9037}$ \\   
                                  & PR     & 0.0021    & 0.0238 & 0.1295 & 0.3719 & 0.4427            & 0.4210 & $\mathbf{0.7839}$ \\   
                                  & F1     & 0.0038    & 0.0187 & 0.1599 & 0.3744 & 0.5172            & 0.3683 & $\mathbf{0.8395}$ \\   
                                  & ACC    & 0.6348    & 0.8358 & 0.9149 & 0.9231 & 0.9381            & 0.9027 & $\mathbf{0.9630}$ \\  \midrule[1.5pt] 
    \end{tabular*}
  \end{table}

\section{Experiments}
\subsection{ Settings}
\subsubsection{Data.}
To better demonstrate the practical value of the proposed model, we conducted sufficient experiments on five datasets of various scales:  Digg\footnote{https://archive.org/details/stackexchange},  Memetracker\footnote{http://snap.stanford.edu/memetracker}, Android\footnote{https://github.com/aravindsankar28/InfVAE/tree/master/data\label{foot1}}, Christianity\textsuperscript{\ref {foot1}}, Twitter\footnote{https://github.com/albertyang33/FOREST/tree/master/data\label{foot2}}. Each dataset includes real information cascade data, and the specific dataset details are shown in the Appendix. In order to unify the training standard with the previous method, we define the nodes at the first 5\% of the infection time in a cascade as dissemination sources, and all the  nodes in the cascade as disseminated observations. We set the ratio of training, validation, and testing to 2:2:6.
\subsubsection{Baselines \& Metrics.}
We adopt the following two types of baselines to make a more comprehensive comparison (please refer to the Appendix  for details). 
(1) Methods of presupposing dissemination mode: NetSleuth \cite{NetSleuth},
OJC \cite{OJC}.
(2) Methods compatible with multiple modes of dissemination:
LPSI \cite{LPSI},
GCNSI \cite{GCNSI},
IVGD \cite{IVGD},
SL-VAE \cite{vaeSL2022}.

Four evaluation matrices are used in the experiments to expose model performance more objectively and comprehensively. First, we use the most commonly used accuracy (ACC), the proportion of correctly classified samples to the total sample. However, since the source localization is essentially an unbalanced classification problem, we added a more appropriate metric, F1-Score (F1) \footnote{$F1=\frac{2PR*RE}{PR+RE}$ } to reconcile the average of precision (PR) and recall (RE). We also attached the results of PR and RE for reference.

\subsubsection{Implementation Details.}
For our SL-Diff model, the details of its implementation are as follows. For the diffusion steps of the coarse stage and the fine stage, we set them to $T_1=800$ and $T_2=80$, respectively. For the specific dissemination model $ f_{\mathbf{w_2}}\left(g_{\mathbf{w_1}}(\mathbf{i})\right)$ we chose DeepIS \cite{deepis}. For the score function approximating network (Equation \ref{equ:approximation}), four layers of MLP are applied.
For other benchmarks, we follow the original structural design. Specifically, for the dissemination model that SL-VAE combines, we adopt DeepIS, while the IC function is used for IVGD.

The models are trained on a single NVIDIA GeForce RTX 3090 GPU.  We use a grid search to find the most appropriate combination of parameters for each model. Specifically, the search range for the number of GCN stacks is from 2 to 8. The learning rate is tuned within $\{5\times10^{-2}, 10^{-2}, 5\times10^{-3}, 10^{-3}\}$. The range of Riemannian SGD weight decay is $\{10^{-2}, 10^{-3}, 10^{-4}, 10^{-5}\}$.

\subsection{Overall Performance}
We have conducted in-depth comparisons of SL-Diff with state-of-the-art baselines on five real datasets, and the results are presented in Table \ref{tb:all}. Generally, models with preset dissemination modes perform relatively poorly, as real information dissemination modes are complex and full of uncertainties. SL-Diff, on the other hand, achieves significantly optimal results in most cases. Among them, the improvement on the Twitter  and Android datasets is particularly significant which can be attributed to its ability to denoise the dissemination noise.  And this noise is more prominent on datasets with a large cascade length and a large number of nodes.
Correspondingly, on the Christianity dataset with a small number of nodes, the performance of SL-Diff is limited. In summary, SL-Diff is more capable of handling large-scale datasets.

\begin{table}[t]
  \centering \scriptsize
  \caption{Ablation study (\textbf{---} represents GPU memory overflow).}
\begin{tabular*}{\textwidth}{@{\extracolsep{\fill}}c|cc|cc|cc|cc|cc}
  \midrule[1.5pt] Datasets & \multicolumn{2}{|c}{ Digg } & \multicolumn{2}{c}{ Memetracker } & \multicolumn{2}{c}{ Android } & \multicolumn{2}{c}{ Christianity } & \multicolumn{2}{c}{ Twitter } \\
  \midrule[0.3pt]& F1 & ACC & F1 & ACC & F1 & ACC & F1 & ACC & F1 & ACC \\
  \midrule[0.3pt] 
  SL-Diff(1) &0.4224 &0.8510   &0.4001 &0.7410   &0.5188 &0.7309   &0.4511 &0.7792   &0.6530 &0.8630 \\
  SL-Diff(2) &\textbf{---} &\textbf{---}   &\textbf{---} &\textbf{---}   &\textbf{---} &\textbf{---}   &0.6024 &0.9530   &\textbf{---} &\textbf{---} \\
  SL-Diff(3) &0.6202 &0.9627   &0.4346 &0.9023   &0.6427 &0.9705   &0.6210 &0.9722   &0.7401 &0.9328 \\
  SL-Diff(4) &0.5985 &0.9329   &0.5792 &0.9527   &0.6132 &0.9595   &0.6285 &0.9717   &0.7620 &0.9441 \\
  SL-Diff    &0.6683 &0.9824   &0.5607 &0.9565   &0.6914 &0.9937   &0.6348 &0.9818   &0.8395 &0.9630 \\
  \midrule[1.5pt]
  \end{tabular*}\label{tb:ablation}
\end{table}

\subsection{Ablation Study}
To verify the effectiveness of each component of our proposed model, we conduct the following ablation experiments. Compared with the complete model, SL-Diff(1) only retains the coarse stage and is initialized randomly.   SL-Diff(2) only retains the fine stage and we take the top 5\% of nodes with the largest source proximity degree generated as the source of prediction. SL-Diff(3) is the model with the cascade positional representations removed, and SL-Diff(4) removes the simulated dissemination noise term in Equation \ref{equ:noise}. The diffusion step of the above four models is all set to 800.
From the results in Table \ref{tb:ablation}, we can conclude that only the coarse stage model can make rough predictions to a certain extent, and it is difficult to make accurate predictions. The model with only the fine stage is not necessarily more accurate than the two-stage model, and the computational overhead is huge, which is not feasible. From the results of SL-Diff(3), we can see that cascade positional representations generally have a more important influence on larger-scale graphs, which may be due to the fact that the relative position of cascades in large-scale graphs will be more complicated, and their influence on propagation will be larger. From the results of SL-Diff(4), we can find that simulated dissemination noise improves the model with a longer cascade length more significantly, which may be because a cascade with a longer propagation chain will introduce relatively more noise.

\begin{figure}[t] 
  \centering

\begin{minipage}[b]{0.32\linewidth}
    \includegraphics[width=\linewidth]{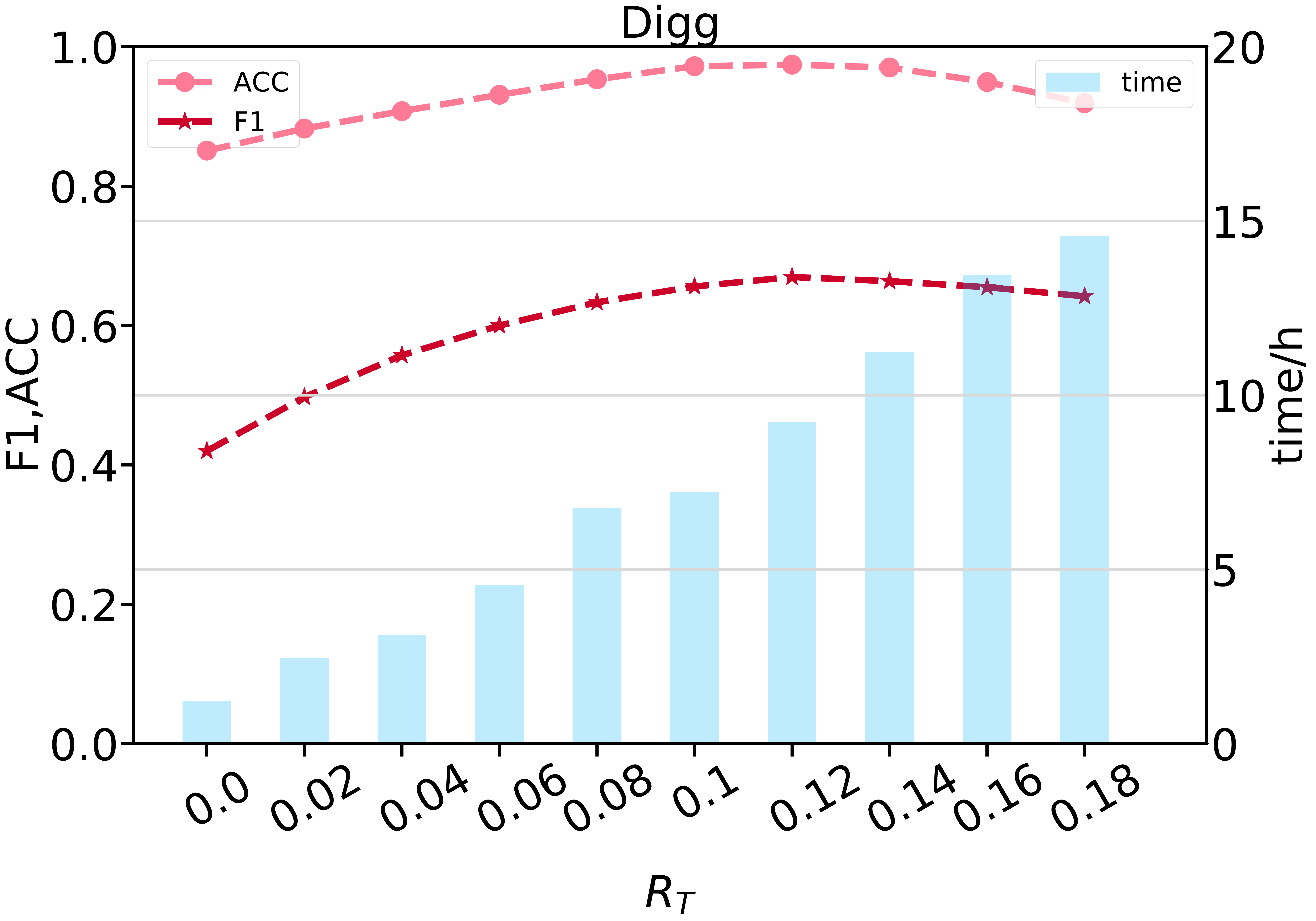}
    \subcaption{}
\end{minipage}
\begin{minipage}[b]{0.32\linewidth}   
    \includegraphics[width=\linewidth]{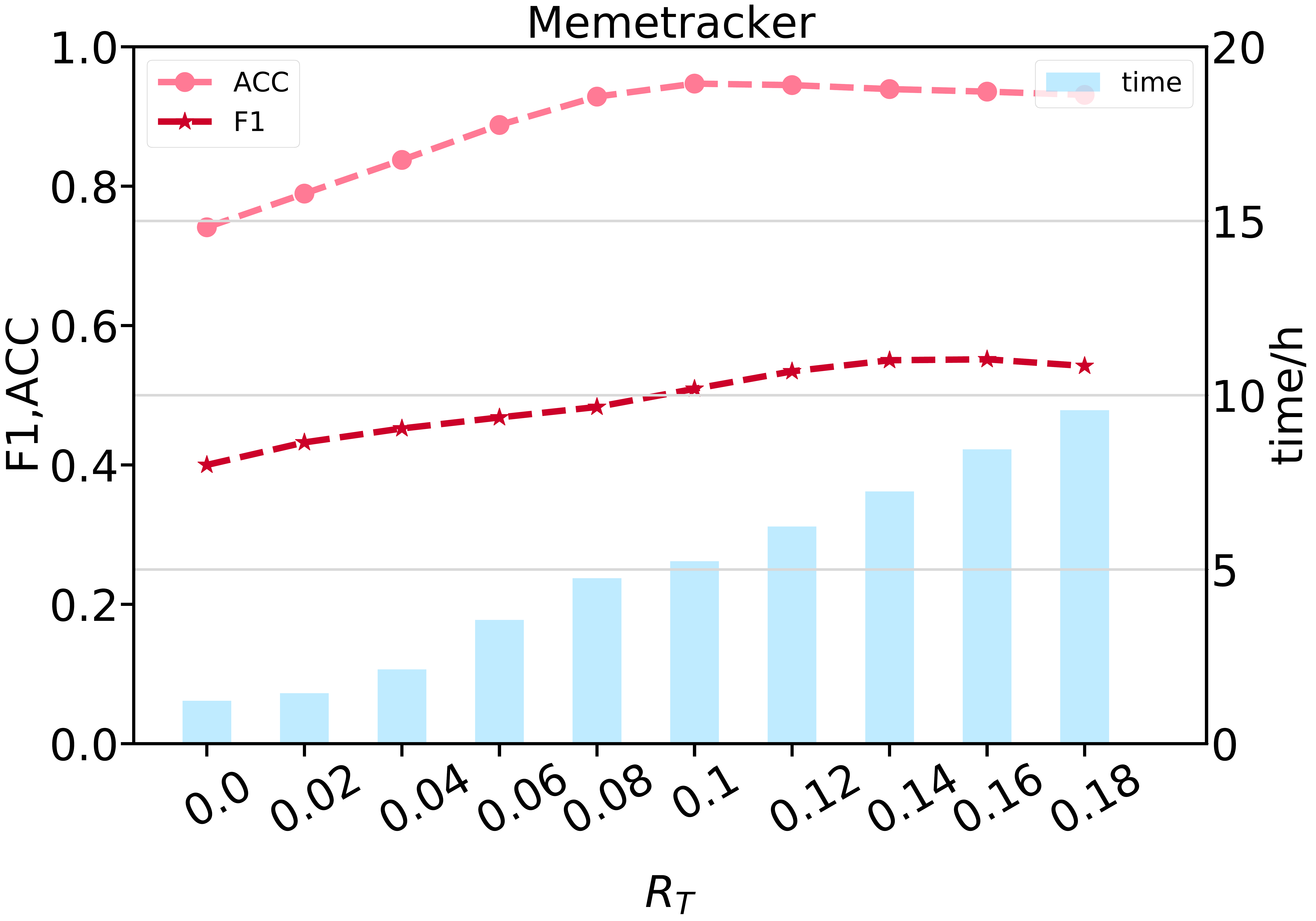}
    \subcaption{}
\end{minipage}
\begin{minipage}[b]{0.32\linewidth}
  \includegraphics[width=\linewidth]{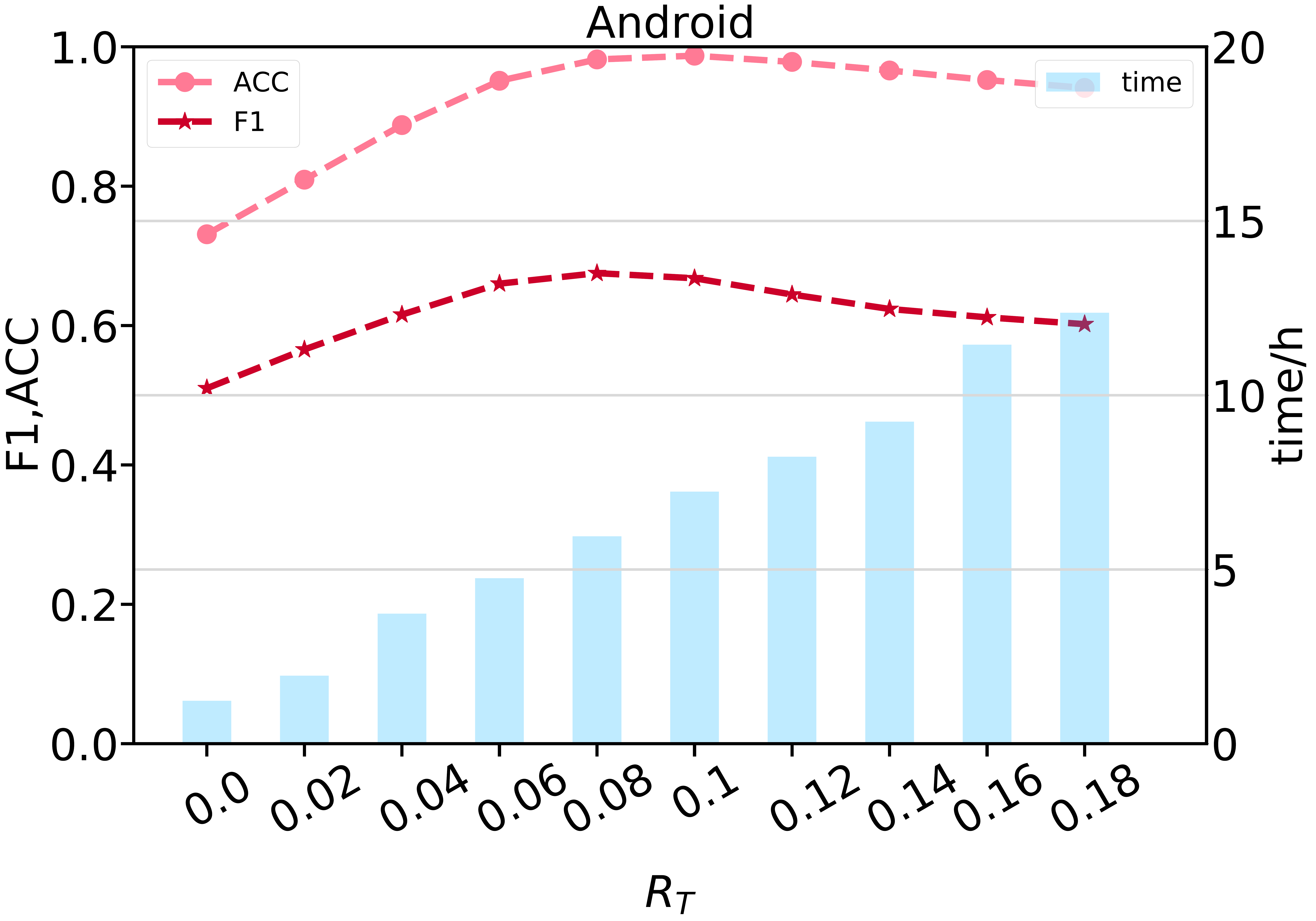}
  \subcaption{}
\end{minipage}
\begin{minipage}[b]{0.32\linewidth}
  \includegraphics[width=\linewidth]{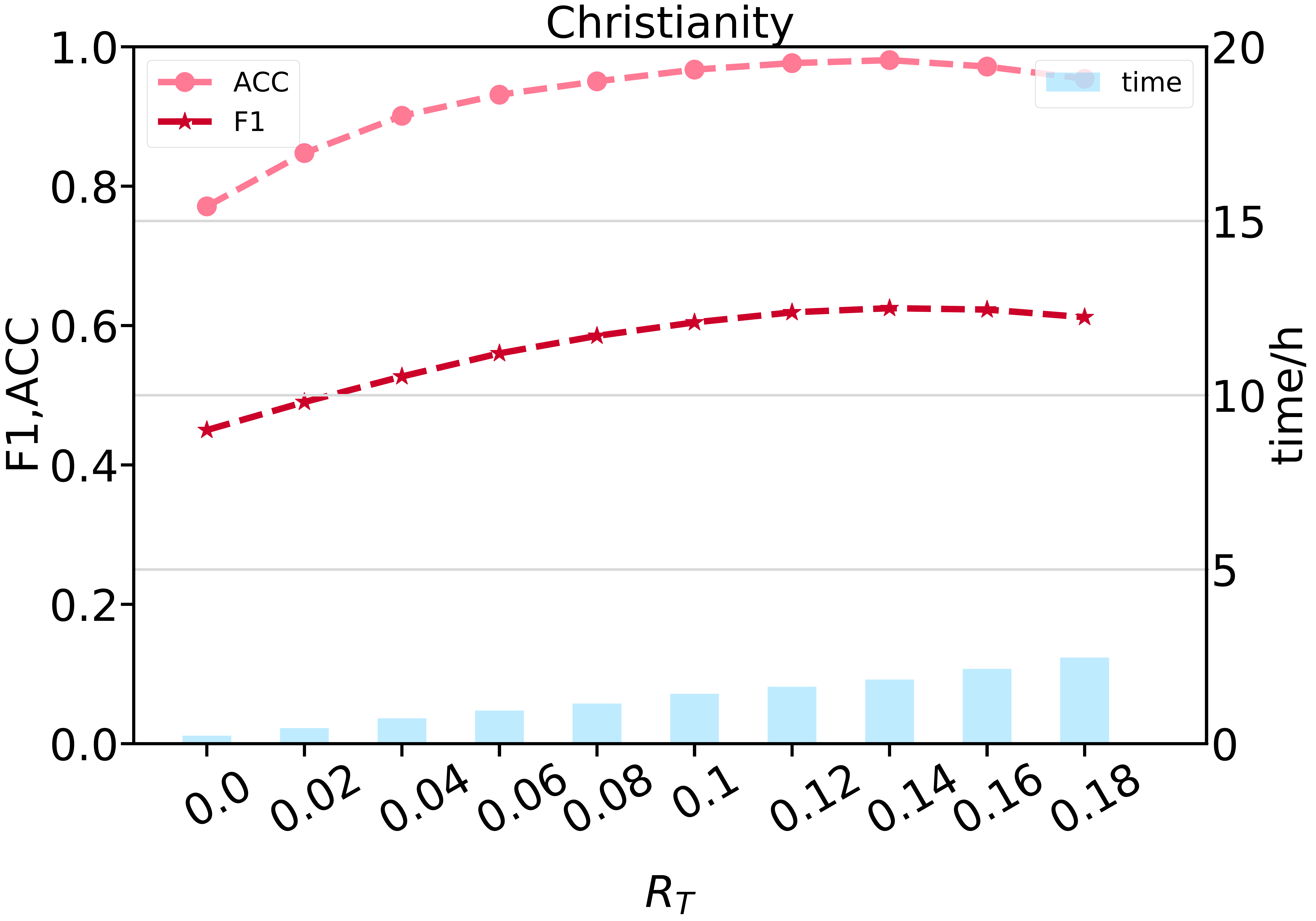}
  \subcaption{}
\end{minipage}
\begin{minipage}[b]{0.32\linewidth}
  \includegraphics[width=\linewidth]{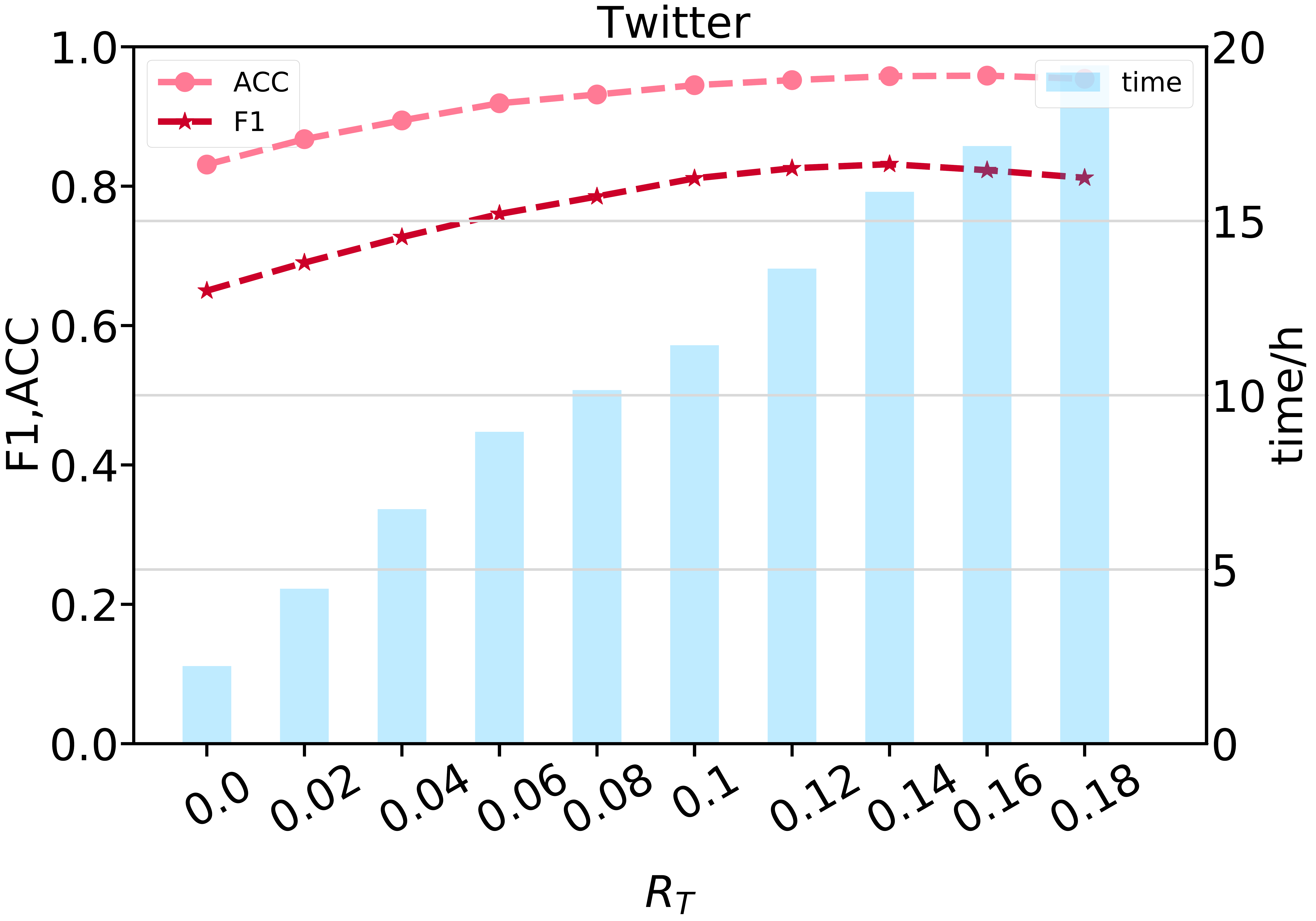}
  \subcaption{}
\end{minipage}
\caption{Analysis on the ratio of the diffusion step of the fine stage to the coarse stage.}
\label{fig:ratio}
 \end{figure}

 \begin{figure}[t] 
  \centering

\begin{minipage}[b]{0.19\linewidth}
    \includegraphics[width=\linewidth]{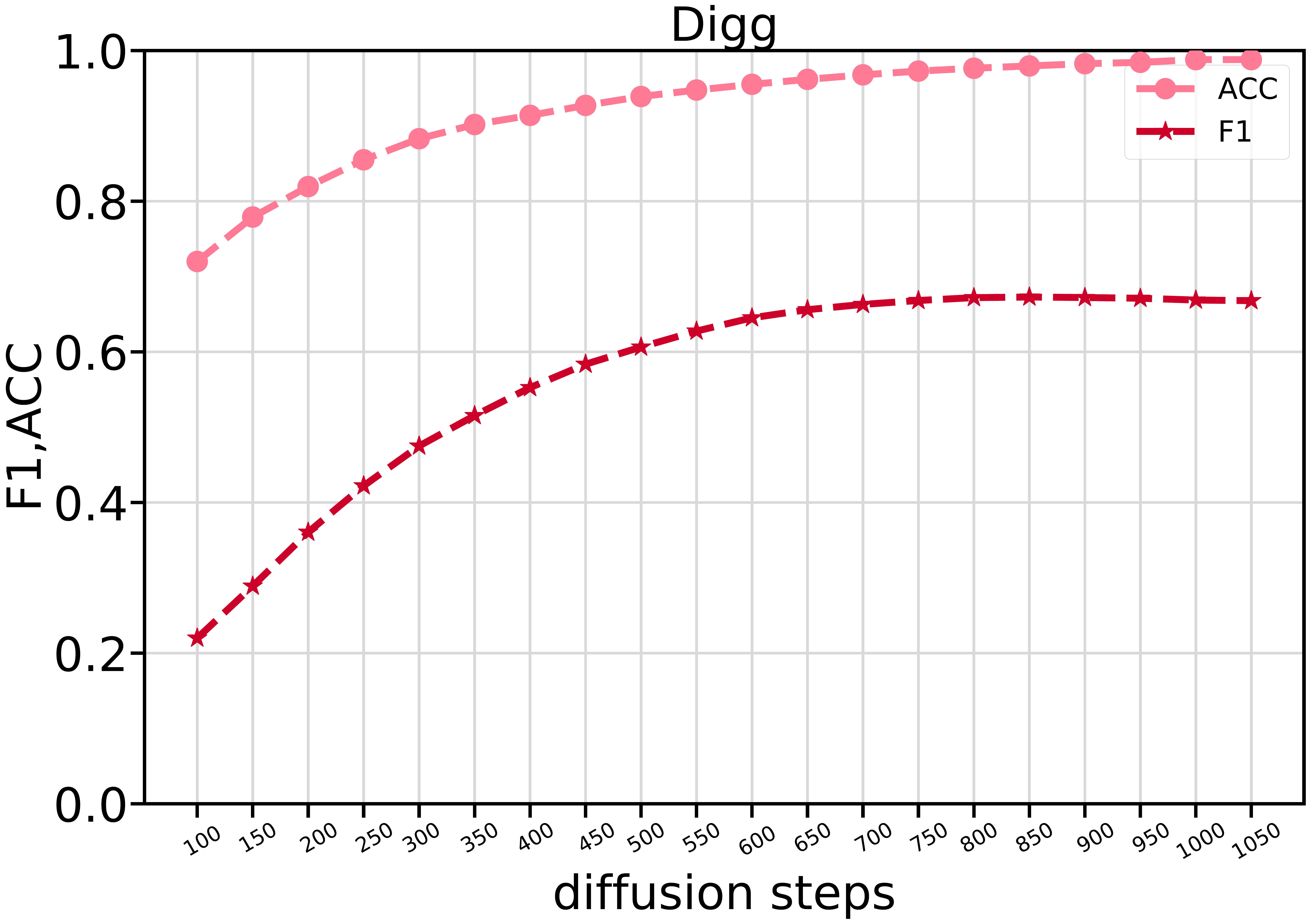}
    \subcaption{}
\end{minipage}
\begin{minipage}[b]{0.19\linewidth}   
    \includegraphics[width=\linewidth]{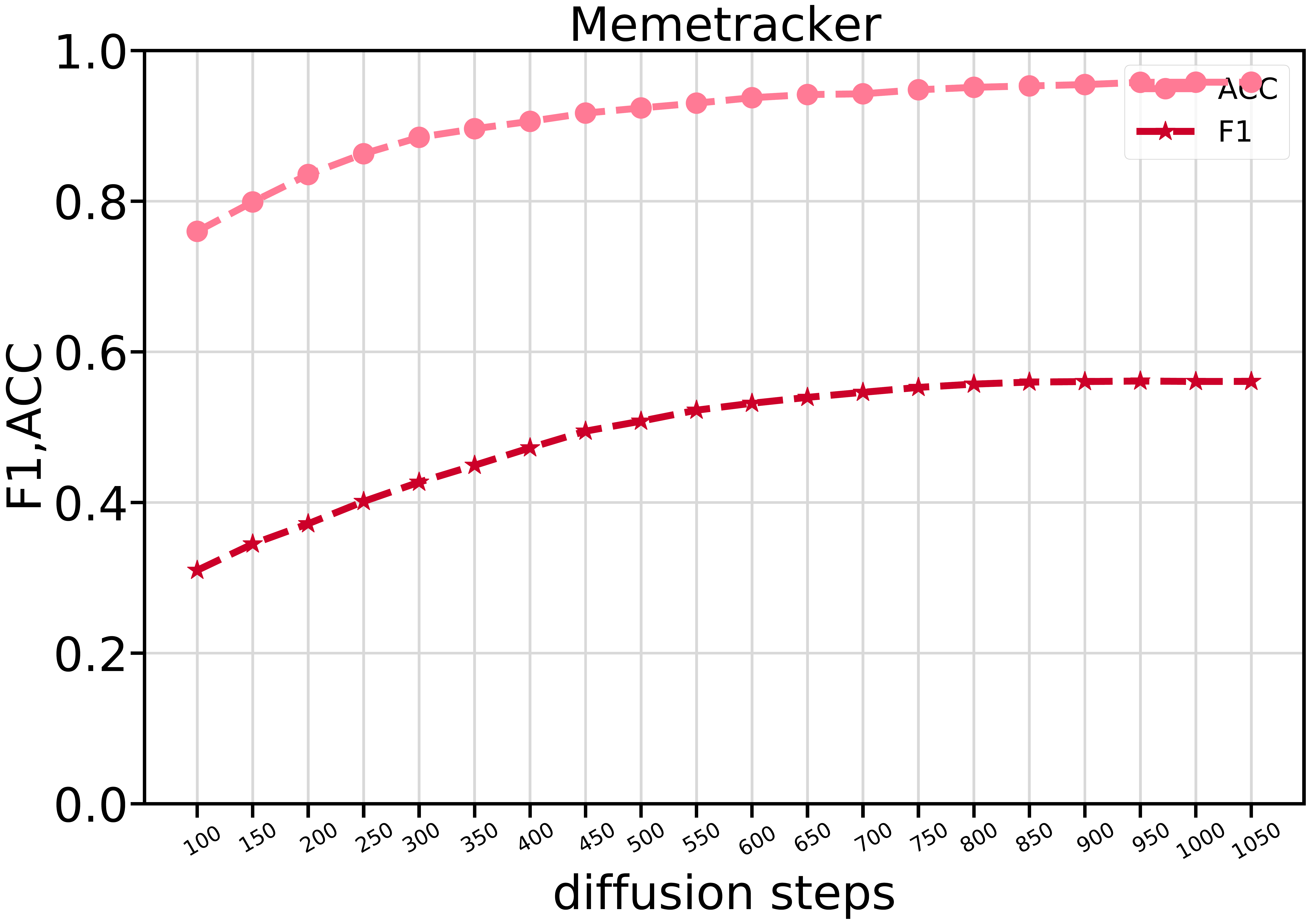}
    \subcaption{}
\end{minipage}
\begin{minipage}[b]{0.19\linewidth}
  \includegraphics[width=\linewidth]{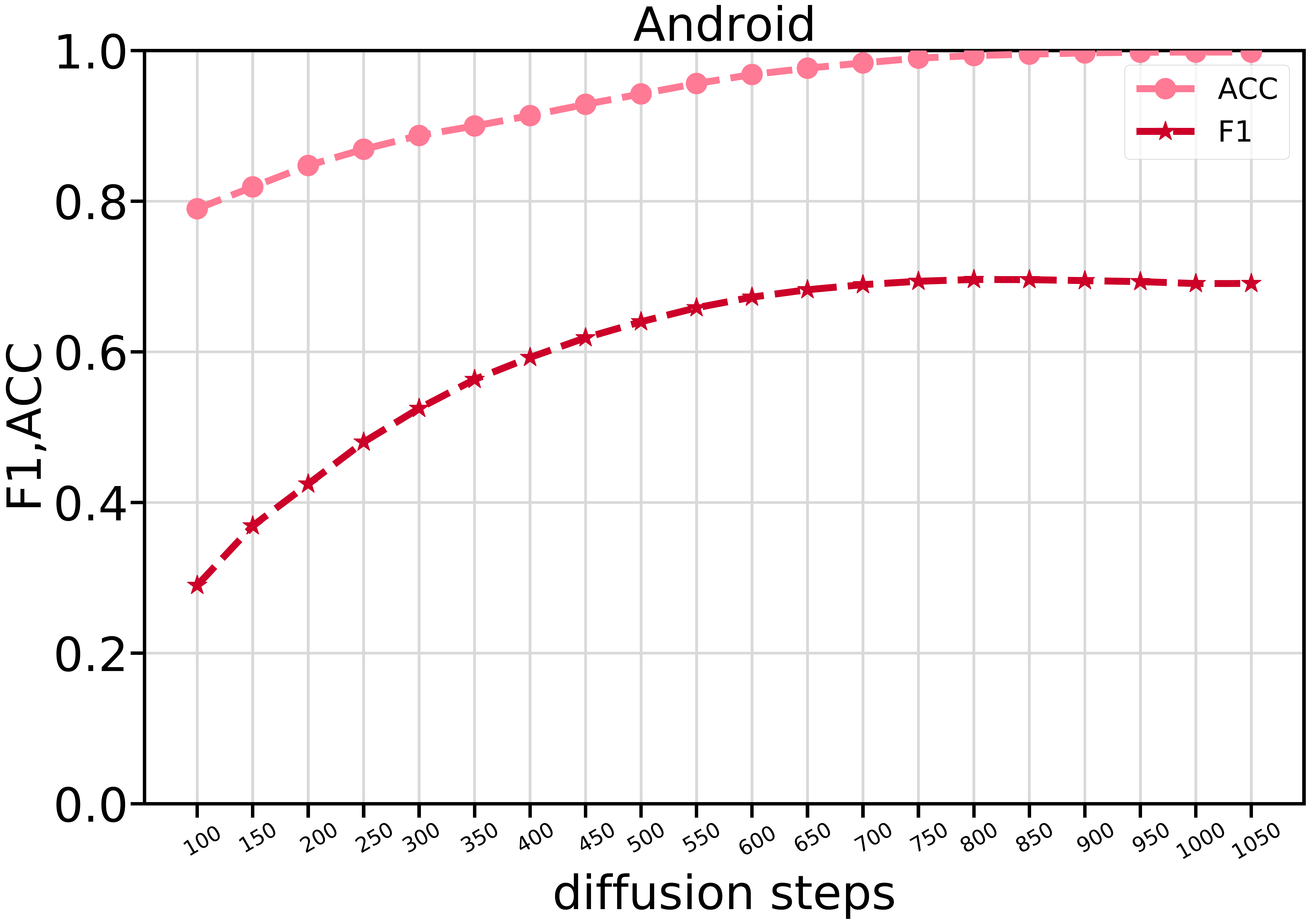}
  \subcaption{}
\end{minipage}
\begin{minipage}[b]{0.19\linewidth}
  \includegraphics[width=\linewidth]{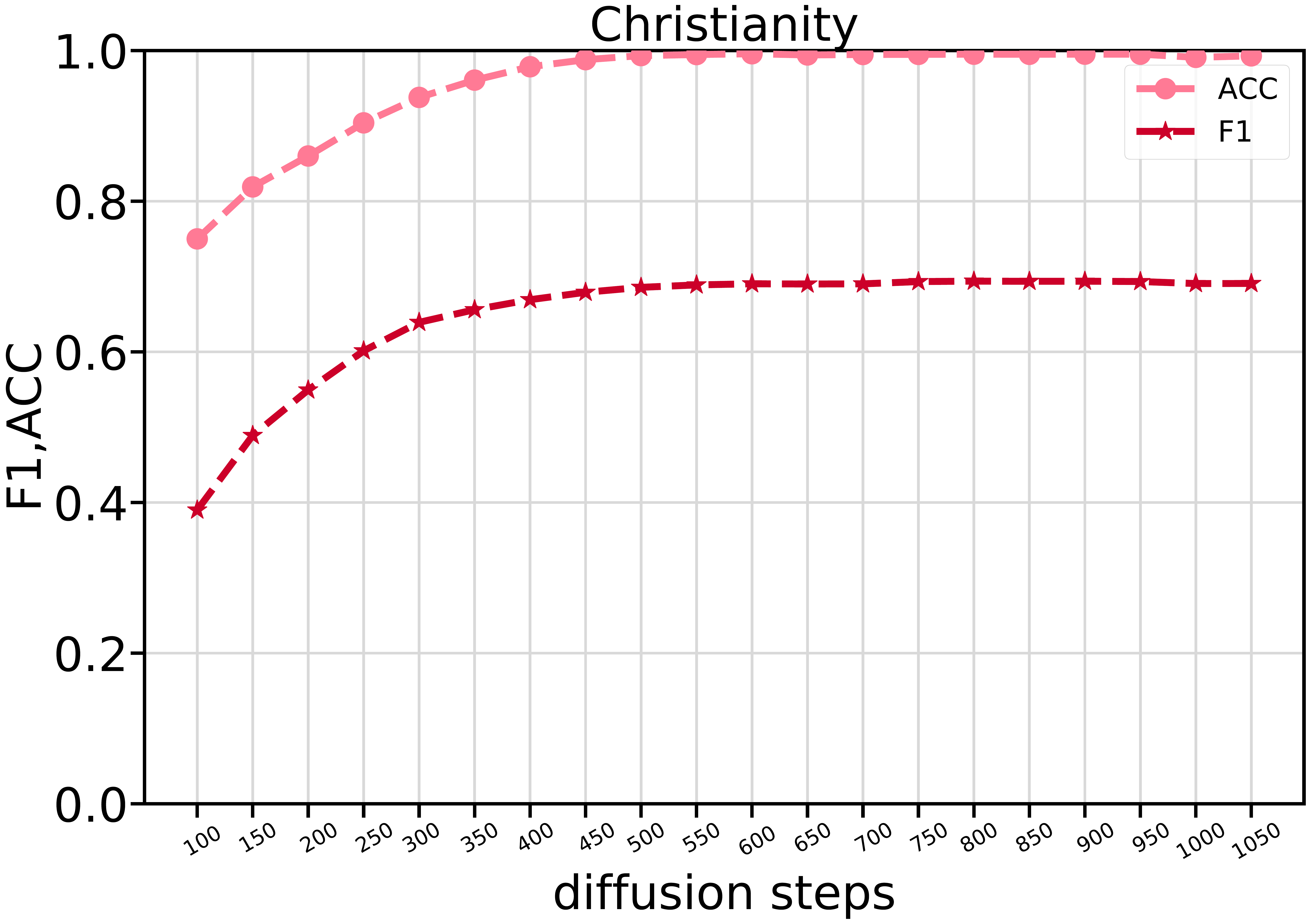}
  \subcaption{}
\end{minipage}
\begin{minipage}[b]{0.19\linewidth}
  \includegraphics[width=\linewidth]{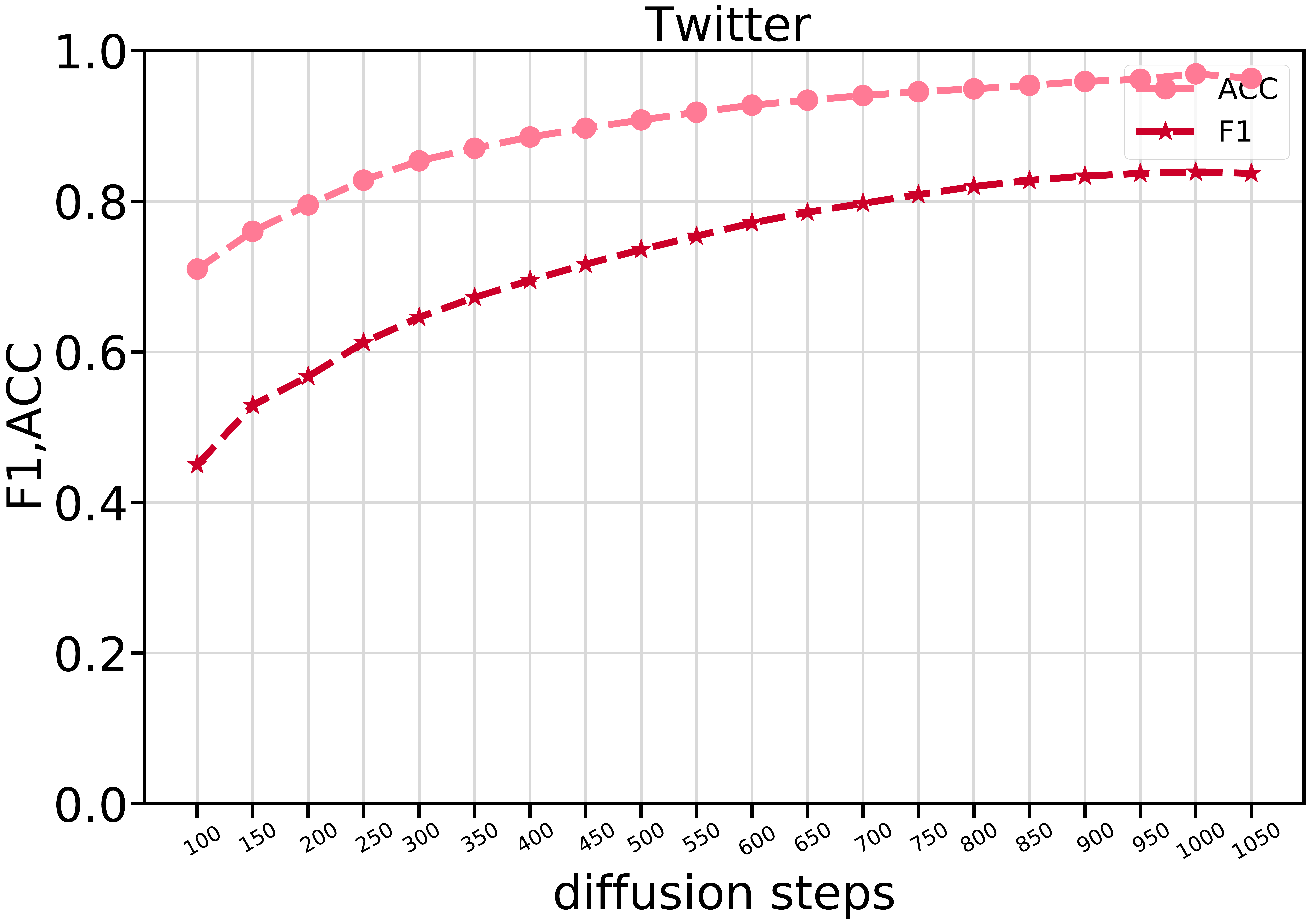}
  \subcaption{}
\end{minipage}
\caption{Convergence analysis.}
\label{fig:steps}
 \end{figure}

\subsection{Parameter Analysis}
In this section, we explore the interdependence of the two stages in greater detail. Specifically, we set the ratio of the diffusion step of the fine stage to the coarse stage as $R_T=\frac{T_2}{T_1}$ and keep the total diffusion step at 800. By adjusting the value of $R_T$, we gain a deeper understanding of how the two stages mutually reinforce each other, and determine the optimal ratio. As shown in Figure \ref{fig:ratio}, we observe that the model's performance initially improves significantly with an increasing proportion of the fine stage, but then gradually declines, while the sampling time steadily increases. This suggests that we can identify the most suitable ratio of the two stages at a relatively low time cost across different datasets. The performance degradation resulting from too high $R_T$ may be due to insufficient denoising in the coarse stage.
Additionally, we analyze the impact of the diffusion step on the model's performance under the configuration with the optimal $R_T$. Figure \ref{fig:steps} reveals that the convergence of the sampling process is comparatively faster for datasets with smaller scales.

\subsubsection{Efficient Analysis.}
\begin{figure}
\centering
\includegraphics[width=1\linewidth]{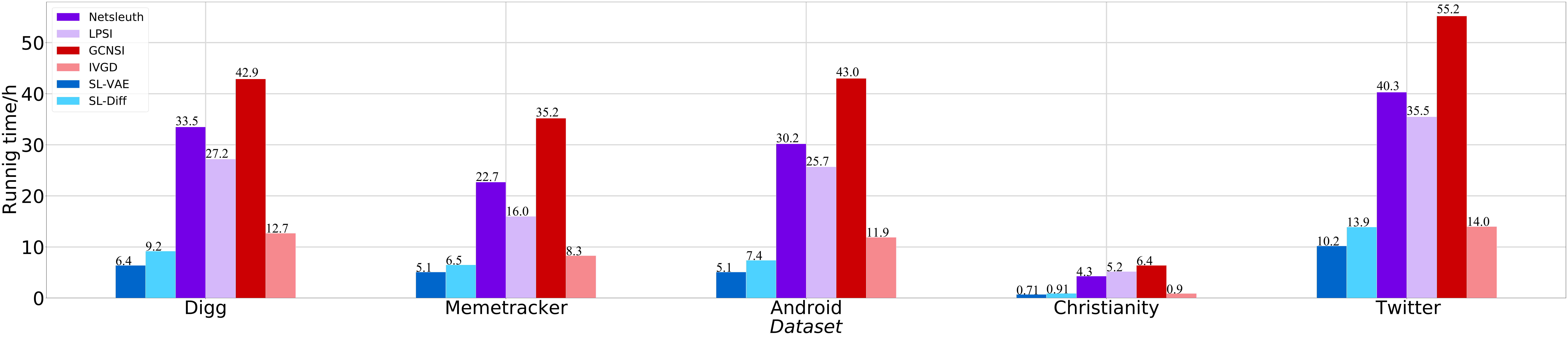}
  \caption{Model efficiency comparison.}
  \label{fig:efficient}
\end{figure}

We compared the training time of various models (sampling time is also included since SL-Diff and SL-VAE are generative models). As shown in Figure \ref{fig:efficient}, the generative model exhibits a shorter overall running time for the source localization task. Among all the models evaluated, SL-Diff effectively controls the computational cost at a lower level, emphasizing the significance of the coarse stage for initialization.

\section{Conclusion}
In this paper, we propose a new two-stage training paradigm to solve the ill-posed problem in graph source localization from the root. SL-Diff also overcomes the difficulties of introducing the powerful diffusion model into this problem, and achieves optimal results on various real-world datasets. The modeling of dissemination noise further improves the approximation performance of SL-Diff. Overall, by leveraging the diffusion model, we have gained deeper insights into the mechanism of the source localization problem, thereby elevating research in this field to a new level.
\bibliographystyle{splncs04}


\end{document}